# Can Large Language Models Replace Human Subjects? A Large-Scale Replication of Scenario-Based Experiments in Psychology and Management

Ziyan Cui[1]

Ning Li[1*]

Huaikang Zhou[1]

[1] Tsinghua University, Beijing, China; [*]E-mail: lining@sem.tsinghua.edu.cn

Abstract

Artificial Intelligence (AI) is increasingly being integrated into scientific research, particularly in the social sciences, where understanding human behavior is critical. Large Language Models (LLMs) have shown promise in replicating human-like responses in various psychological experiments. We conducted a large-scale study replicating 156 psychological experiments from top social science journals using three state-of-the-art LLMs (GPT-4, Claude 3.5 Sonnet, and DeepSeek v3). Our results reveal that while LLMs demonstrate high replication rates for main effects (73-81%) and moderate to strong success with interaction effects (46-63%), They consistently produce larger effect sizes than human studies, with Fisher Z values approximately 2-3 times higher than human studies. Notably, LLMs show significantly lower replication rates for studies involving socially sensitive topics such as race, gender and ethics. When original studies reported null findings, LLMs produced significant results at remarkably high rates (68-83%) - while this could reflect cleaner data with less noise, as evidenced by narrower confidence intervals, it also suggests potential risks of effect size overestimation. Our results demonstrate both the promise and challenges of LLMs in psychological research, offering efficient tools for pilot testing and rapid hypothesis validation while enriching rather than replacing traditional human subject studies, yet requiring more nuanced interpretation and human validation for complex social phenomena and culturally sensitive research questions.

Artificial intelligence (AI) is rapidly transforming scientific research, a shift often referred to as “AI for science” (1,2). In the social sciences, where understanding human behavior, cognition, and perception is key, large language models (LLMs) are emerging as powerful tools that could reshape established research methods (3–5), which have long relied on experiments, surveys, and interviews with human participants.

Early studies have shown mixed results in using LLMs to replicate human responses in psychological assessments (6–9) and economic decision-making (10–14). While some research demonstrates close alignment between LLM-generated and human responses (6,8,15), other studies reveal notable divergences, particularly in simulating individual-level behaviors and specific demographic profiles (16-19). These contrasting findings raise important questions about the reliability and applicability of LLMs across different research contexts. Moreover, existing studies often rely on a limited number of arbitrarily selected experiments, making it difficult to draw comprehensive conclusions about LLMs’ capabilities across diverse psychological phenomena and experimental conditions (3,5).

Crucial questions remain: To what extent can LLMs supplement or even replace human subjects across diverse psychological experiments? Are there systematic differences between human and AI responses, particularly in areas where such divergences might be more pronounced, such as socially sensitive topics (20)? Addressing these questions is vital for determining the applicability and limitations of LLMs in social science research (3).

To fill this critical gap, we conducted a large-scale study replicating 156 randomly selected psychological experiments from five top management and psychology journals. In these replications, we presented the original experimental materials to three different advanced LLMs—GPT-4, Claude 3.5 Sonnet, and DeepSeek V3—instead of human participants. This approach, known as "silicon replication," treats each LLM response as analogous to a human participant's response. For each original study, we generated an equivalent number of LLM responses to ensure balanced comparisons across all experiments. We focused on text-based vignette studies, a method commonly used in organizational and general psychology (21,22) and featured in Nobel Prize-winning work on decision-making (23–25). These experiments are suitable for LLM replication as they rely on participants responding to textual stimuli with decisions, choices, and cognitive expressions.

We systematically evaluate LLMs' capabilities across diverse management and psychological topics, including socially sensitive areas such as race, gender, and ethical scenarios. Using standard replication indicators—replication rates, p-value distributions, effect sizes, and study feature influences (26–28)—our analysis reveals that LLMs' application in psychological science presents both promising capabilities and notable challenges.

Across the three LLMs, we found consistently high replication rates spanning different journals, samples, study types, and research topics. Nearly three-quarters of main effects and approximately half of interaction effects were successfully replicated, with the direction and statistical significance of original findings preserved. For instance, GPT-4 achieved replication rates of 72.7% for main effects and 45.7% for interaction effects, with the other two LLMs demonstrating even higher replication rates across both types of effects. The lower success rate in replicating interaction effects aligns with known challenges in human participant studies, where such effects are typically more difficult to detect (29,30). These robust replication rates suggest LLMs could meaningfully contribute to psychological research by offering a scalable tool for theory testing and validation.

Despite the general success, we found that studies involving socially sensitive topics, such as race and gender, were significantly less likely to be replicated, though the magnitude of this effect varied across models. For example, GPT-4's main effect replication rate dropped dramatically from 76.8% in studies without race variables to 41.5% in studies with race variables. This stark difference may be attributed to LLMs' alignment with certain values and their tendency to respond in socially desirable ways, even in hypothetical situations (31,32). These value alignments appear to make LLMs more cautious and less prone to producing responses that could be considered controversial (33,34), potentially compromising their ability to accurately replicate studies where social sensitivity plays a significant role.

Beyond replication rates, we observed systematic differences in effect sizes between original studies and LLM replications. Effect sizes generated by LLMs were consistently larger than those in the original human studies, suggesting potential effect amplification, with Fisher Z values approximately 2-3 times higher than human studies. Notably, LLMs produced significant main effects in 68-83% of cases where original studies reported null findings. This pattern across

LLMs may reflect their unique characteristics as research subjects: unlike human participants, LLMs operate without fatigue, distraction, or response inconsistency, potentially allowing them to detect subtle psychological effects that might be obscured by human response variability. However, this heightened sensitivity presents a dual interpretation—while it might help identify previously undetectable patterns in psychological phenomena, it could also indicate a tendency toward Type I errors.

These findings reveal both the potential and inherent complexities of using LLMs in experiment research. While LLMs demonstrate promise as simulated agents for piloting studies, testing instruments, and exploring theoretical mechanisms (35,36), their distinct response patterns—including heightened sensitivity to subtle effects and potential for effect size amplification—underscore the need for nuanced implementation. We suggest treating LLMs not as substitutes for human participants, but as complementary research tools that, when properly calibrated and interpreted, can enhance our understanding of human psychology and behavior while opening new avenues for methodological innovation.

## Methods

### Study Overview

We aim to systematically evaluate whether three leading Large Language Models (LLMs) - GPT, Claude, and DeepSeek[1] - can replicate human responses in psychological experiments,

---

[1] We also tested two additional open-source LLMs (Llama 3.1 and Mistral-Large), but excluded them from the main analysis due to consistent formatting and completion issues. For instance, when prompted to generate 20 item responses, models often produced only 5-6 items; in samples requiring 50 responses, they frequently generated empty outputs with only 3-4 usable cases, or produced responses entirely irrelevant to the experimental prompts.

employing a carefully designed procedural framework (Figure 1).

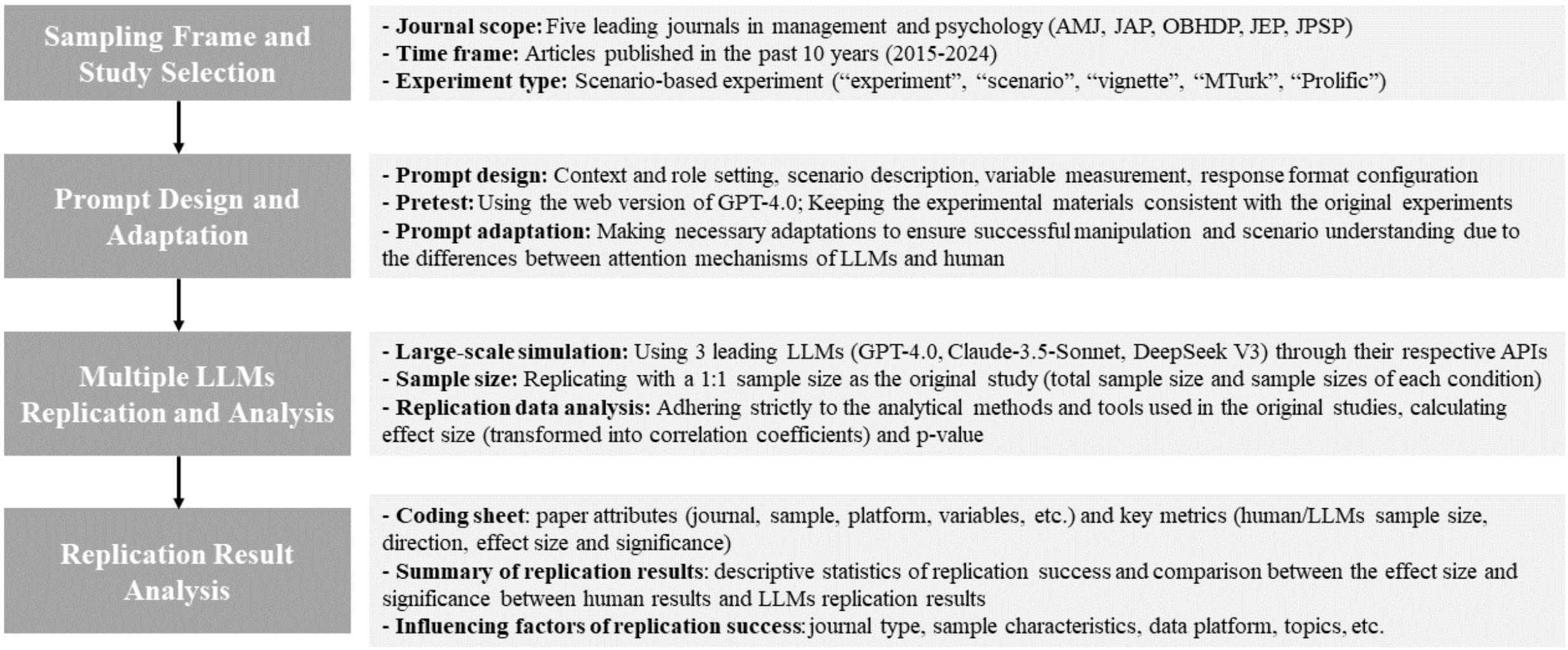

**Figure 1: Research Design and Process**

We focused on scenario-based experiments published between 2015 and 2024 in five leading social science journals[2]: three from management (*Organizational Behavior and Human Decision Processes, Academy of Management Journal, Journal of Applied Psychology*) and two from psychology (*Journal of Personality and Social Psychology, Journal of Experimental Psychology: General*). Using Google Scholar, we identified relevant articles using keywords such as "experiment," "scenario," "vignette," "MTurk," and "Prolific."

From this pool, we randomly selected 10 articles from each journal, applying specific exclusion criteria to ensure LLM replication feasibility. We excluded studies involving self-reported real-life experiences, priming techniques for motivation/emotion/cognition, physiological measurements or behavioral observations, longitudinal designs, and team/group interactions. These exclusions were necessary as LLMs cannot draw on personal experiences, physically act, or replicate the nuanced human interactions and long-term processes these studies require. Articles not meeting these criteria were replaced through random selection until reaching 10 suitable articles per journal.

[2] The use of experiments published between 2015 and 2024 does not raise significant concerns about LLMs such as GPT-4 being trained on these studies, as its training data typically excludes full academic manuscripts from subscription-based journals. Our preliminary tests showed that GPT-4 consistently failed to provide accurate information about these articles. Furthermore, our design only exposed LLMs to one experimental condition at a time, keeping them blind to the overall study design and hypotheses. We present additional analyses in later sections to further rule out this possibility.

Our final sample consisted of 156 studies from the 50 selected articles, with a balanced representation across disciplines: 70 studies from management journals and 86 studies from psychology journals. The dataset for final analysis of GPT-4 replication contained 690 main effects and 164 interaction effects[3]. Analysis of effects per study showed consistency between management (4.37 effects/study) and psychology (4.47 effects/study) journals for main effects, with no significant differences between disciplines (Kruskal-Wallis test: H = 0.029, $p = 0.865$). The same consistency was observed for interaction effects (management: 2.36, psychology: 2.14 effects/study; H = 0.147, $p = 0.702$). Similar patterns were observed for Claude and DeepSeek[4].

## Data and Code Availability Statement

[3] Due to DeepSeek's model limitations in processing images, 64 main effects and 33 interaction effects involving visual stimuli were not included in DeepSeek's replication attempts. For Claude, the dataset contained 683 main effects and 164 interaction effects. For DeepSeek, the dataset contained 618 main effects and 131 interaction effects. Details see Method Details section of the SI.

[4] For Claude, analysis of effects per study showed consistency between management (4.33 effects/study) and psychology (4.47 effects/study) journals for main effects, with no significant differences between disciplines (Kruskal-Wallis test: H = 0.176, $p = 0.675$). The same consistency was observed for interaction effects (management: 2.36, psychology: 2.14 effects/study; H = 0.147, $p = 0.702$). For DeepSeek, analysis of effects per study showed consistency between management (4.52 effects/study) and psychology (4.26 effects/study) journals for main effects, with no significant differences between disciplines (Kruskal-Wallis test: H = 0.323, $p = 0.570$). The same consistency was observed for interaction effects (management: 2.36, psychology: 1.64 effects/study; H = 2.568, $p = 0.109$).

Essential information is provided in the main text, with detailed methods and additional analyses available in the supplementary information (SI). All research materials, including the complete dataset, experimental prompts, code used for API calls, and coding sheet, are accessible via the Open Science Framework (OSF) repository at: https://osf.io/j6wmn/?view_only=5947919c57a440ddb02e5e07ac069a5f.

**Prompt Design and Adaptation**

We structured each prompt into four key components: (1) context and role setting, (2) scenario description, (3) variable measurement, and (4) response format configuration. The context setting established the experimental frame, specifying participant role and task requirements ("Imagine you are a person invited to participate in an experiment"). Scenario descriptions preserved original experimental materials, while variable measurement guided the model in assessing key study outcomes. All responses were configured in JSON format to enable systematic analysis.

Before large-scale data collection, we conducted comprehensive prompt validation testing, similar to pilot studies in human experiments. For each experimental condition, we tested prompts multiple times to evaluate whether LLMs consistently detected manipulations and produced interpretable responses aligned with the study's variables of interest. When LLMs exhibited "failed attention" - such as ignoring critical framing elements or missing essential context - we restructured prompts to highlight these aspects, similar to how researchers might emphasize critical priming information for human participants. Through this validation process, we found that 35.3% of studies required prompt adaptations to ensure reliable engagement with experimental manipulations.

While these adaptations occasionally shifted us from strict literal replications toward more conceptual ones, they were necessary to ensure LLMs accurately processed the experimental manipulations as intended. We systematically coded these adaptations to analyze their impact on replication outcomes, maintaining transparency about where and why modifications were needed (details see Method Details section of the SI). Importantly, our analysis in Figure 2 also showed that studies without prompt adaptations achieved comparable replication rates, suggesting that these modifications did not artificially drive successful replications.

**LLM Replication Process and Data Analysis**

We conducted large-scale simulations using three prominent LLMs - GPT-4, Claude 3.5 Sonnet, and DeepSeek V3 - through their respective APIs to generate responses for our entire sample of experiments[5]. For each model, we used their default temperature setting of 1.0 (except DeepSeek at 1.3, as recommended for general conversation) to balance response diversity with consistency while maintaining the required JSON output format. This approach ensured reliable data collection while preserving the natural variation in model responses.

[5] It is important to note that our study methodology reflects a GPT-centric approach in several respects. Our prompt development and validation process was initially conducted using GPT-4, with these optimized prompts subsequently applied to Claude and DeepSeek models without model-specific adaptations. This sequential development approach was adopted for practical reasons, as GPT-4 was our primary model of interest when the study began, with the other models added later to provide comparative insights.

We recognize that the fundamental nature of LLM response variation differs substantially from human participant variation. While human samples capture genuine individual differences in psychology, personality, and experience, LLM responses with temperature settings primarily generate statistical variations around a central tendency determined by the model's training. This distinction is crucial for interpreting our results[6].

To replicate the original studies, we maintained a 1:1 sample size match with the original experiments, replicating each condition with exactly the same number of participants. While traditional replication studies often focus on achieving sufficient statistical power to detect effects, we note that power considerations apply differently to LLM-based replications due to their typically more homogeneous response patterns. Although this homogeneity might suggest that fewer responses would be sufficient for detecting effects, our precise matching strategy ensures direct comparability of effects and maintains methodological consistency with the source studies, eliminating potential alternative explanations due to sample size differences.

In our analysis, we adhered strictly to the analytical methods and tools used in the original studies to ensure comparability[7]. Our analyses included a range of statistical techniques, such as descriptive statistics, regression analysis, ANOVA, t-tests, structural equation modeling, and chi-square analysis. When the original study did not specify an analytical method, we employed the most commonly used approaches in the field.

---

[6] In rare cases (less than 3% of replications), LLMs produced uniform responses across the entire sample, resulting in zero variance. These instances made it impossible to calculate certain statistical indicators (e.g., correlation coefficients, standard deviations). Such cases were excluded from our analyses and are documented in detail in online supplementary information.

[7] In our analysis, we prioritized direct computation of Cohen's d or eta-squared from the original data to minimize conversion inaccuracies for effect size comparisons. For some experiments that employed regression-based methods, where regression coefficients could not be directly converted to correlation coefficients, we recalculated the original data (if available) and changed the analysis method to ANOVA or t-tests to obtain Cohen's d or eta-squared values.

**Replication Analysis and Comparison**

We conducted a comprehensive analysis of the replication results across all three LLMs, focusing on the reproducibility of main effects and interaction effects reported in the original articles. We developed a detailed coding sheet for each effect, capturing essential information such as journal, sample characteristics, data collection platforms, variables involved, and key metrics such as sample sizes, *p* values, effect directions, and effect sizes for both human and LLM studies. The coding included categorizing topics into different domains, particularly socially sensitive topics like race, gender, and ethics—areas where LLMs may respond differently (34).

We standardized the direction of effects and converted various reported effect size metrics—such as Cohen's d, eta-square, F-statistics, and chi-square—into correlation coefficients (*r*) for consistency and to facilitate interpretation (26). For LLM replications, if the direction of the effect was opposite to that of the original study, the *r* value was recoded as negative. When original studies only reported *p*-value ranges, we calculated precise *p*-values using the *r* values and sample sizes for required analyses.

Given that main effects and interaction effects may behave differently, we structured our analysis by separating these two distinct types of effects. We created two distinct samples for different analytical purposes. The full sample included all usable effects, regardless of the original findings' statistical significance, which we used for general comparisons between human studies and LLM replications across various experimental conditions. For assessing replication success specifically, we analyzed a subsample of statistically significant original findings ($p$ <.05), as replication success fundamentally involves reproducing previously demonstrated significant effects (28).

Our analysis focused on three key areas: replication success rates, effect sizes, and factors influencing replication outcomes (27,37). First, we examined replication success rates across all three LLMs to evaluate how well each LLM replicated supported effects. Next, we compared effect sizes between the original human studies and LLM simulations to assess whether different models vary in their ability to match human response magnitudes. Finally, we used regression analyses to explore how study attributes—such as journal type, sample characteristics, data collection methods, and topic nature—contribute to heterogeneity in LLM-based replication outcomes.

## Results

Our dataset for GPT-4 replication analysis contained 690 main effects and 164 interaction effects from the original studies[8]. The use of these effects varied across different analyses based on methodological requirements. For p-value comparisons, which examine statistical significance regardless of effect direction, we utilized the complete sample. For effect size comparisons, we focused on the subset of effects with unambiguously stated directions[9], including 606 main effects (87.8% of total main effects) and 110 interaction effects (67.1% of total interaction effects)[10]. Our replication success analysis, requiring both clear directions and original statistical significance (28), concentrated on 454 main effects (65.8% of total main effects) and 81 interaction effects (49.4% of total interaction effects)[11].

[8] For Claude, the dataset contained 683 main effects and 164 interaction effects. For DeepSeek, the dataset contained 618 main effects and 131 interaction effects. Details see Method Details section of the SI.

**Replication Success Rate**

Our analysis examined replication success across three LLMs. GPT-4 achieved replication rates of 72.7% for main effects and 45.7% for interaction effects, with other LLMs showing comparable, though slightly higher, success rates (Claude: 80.6% main effects, 61.7% interaction effects; DeepSeek: 76.2% main effects, 62.9% interaction effects). To systematically examine potential variations in replication success, we analyzed replication rates across different categories and conducted proportion tests to assess whether replication rates significantly differed between these categories. Due to the substantially smaller sample size for interaction effects (N = 81 vs. N = 454 for main effects), which would result in too few cases when breaking down into analytical categories, we focused our statistical analyses primarily on main effects.

---

[9] Direction clarity refers to unambiguous predictions about effect direction (e.g., Group A performing better than Group B, positive correlations between variables), excluding effects from analyses where directionality was ambiguous—such as ANOVA results showing differences among three or more groups without specifying which group should perform best, or interaction effects without clear predictions about the pattern of differences.

[10] For Claude, effects with unambiguously stated directions included 600 main effects (87.8% of total main effects) and 110 interaction effects (67.1% of total interaction effects). For DeepSeek, effects with unambiguously stated directions included 551 main effects (89.2% of total main effects) and 97 interaction effects (74.0% of total interaction effects). Details see Method Details section of the SI.

[11] For Claude, effects with both clear directions and original statistical significance included 448 main effects (65.6% of total main effects) and 81 interaction effects (49.4% of total interaction effects). For DeepSeek, effects with both clear directions and original statistical significance included 411 main effects (66.5% of total main effects) and 70 interaction effects (53.4% of total interaction effects). Details see Method Details section of the SI.

Complete results for interaction effects across all categories are provided in the supplementary materials.

Across all LLMs, we observed several consistent patterns in replication success in Figure 2. Studies from psychology journals showed significantly higher replication rates compared to management journals, with GPT-4 achieving 79.1% for psychology versus 64.3% for management studies (z = 3.502, *p* <.001)[12]. This pattern was not observed across other LLMs (Claude: 83.3% vs 77.0%, z = 1.668, *p* = .095; DeepSeek: 78.8% vs 73.0%, z = 1.370, *p* = .171). This platform effect was uniquely observed in GPT-4, which showed 75.5% success for online platforms compared to 66.4% for other recruitment methods (z = -1.998, *p* = .046), while other models showed no statistically significant differences between recruitment methods (Claude: 79.2% vs 83.7%, z = 1.127, *p* = .260; DeepSeek: 75.4% vs 77.7%, z = .524, *p* = .600). Other methodological variations, including scenario type (text-only vs. picture-included), prompt alterations, and sample type (student vs. non-student participants) did not show significant differences in replication rates[13].

---

[12] For manuscript conciseness, we report test statistics and exact p-values (from two-tailed tests) in the main text, while complete statistical outputs—including effect sizes, confidence intervals, degrees of freedom, and detailed methodological specifications—are available in our OSF repository. Exact p-values are reported throughout except when *p* <.001.

[13] The replication success rates of other methodological variations are shown as below: scenario type (text-only vs. picture-included; GPT-4: 72.1% vs. 80.0%, z = -1.011, *p* = .312; Claude: 80.4% vs. 82.9%, z = -0.355, *p* = .723), prompt alterations (yes vs. no; GPT-4: 68.3% vs. 75.5%, z = 1.688, *p* = .091; Claude: 78.3% vs. 82.1%, z =0.985, *p* = .324; DeepSeek: 73.9% vs. 77.6%, z = 0.849, *p* = .396), and sample type (student vs. non-student participants; GPT-4: 76.5% vs. 72.4%, z = 0.515, *p* = .607; Claude: 85.3% vs. 80.2%, z =.723, *p* = .470; DeepSeek: 81.8% vs. 75.7%, z = 0.796, *p* = .426). And it should be noted that DeepSeek does not possess multimodal capabilities, so it was unable to conduct experiments included picture.

To address concerns about potential training data exposure—where LLMs might better replicate studies they were likely exposed to during training—we conducted three analyses. For GPT-4, we did not find statistical differences between pre- and post-2022 studies (75.0% vs. 68.2%, $p = .123$), open access versus paywalled articles (73.5% vs. 65.1%, $p = .242$), and highly versus less cited papers (72.1% vs. 73.4%, $p = .760$). Similar non-significant patterns were observed for Claude and DeepSeek, suggesting that potential training data exposure did not systematically influence any LLM's ability to replicate research findings[14].

Importantly, we examined replication rates for studies involving socially sensitive topics (25,33–35), revealing distinct patterns across models. GPT-4 showed markedly lower replication rates for race-related effects (41.5% vs. 76.8%, $p <.001$) and gender-related effects (57.8% vs. 74.3%, $p = .018$). DeepSeek demonstrated similar but marginally significant disparities for both gender (60.9% vs. 77.1%, $p = .077$) and race (66.0% vs. 77.7%, $p = .064$). Claude showed a different pattern: with no statistically significant differences for gender (75.0% vs. 81.1%, $p = .350$) or race-related studies (77.8% vs. 81.0%, $p = .579$), though it showed somewhat lower replication rates for studies involving ethical considerations (76.6% vs. 82.7%, $p = .125$), though this difference did not reach statistical significance. These varying patterns suggest that different LLMs handle socially sensitive topics in distinct ways.

[14] For Claude, we did not find statistical differences between pre- and post-2022 studies (81.1% vs. 79.6%, $p = .708$), open access versus paywalled articles (80.0% vs. 86.0%, $p = .341$), and highly versus less cited papers (79.1% vs. 82.2%, $p = .421$). For DeepSeek, we did not find statistical differences between pre- and post-2022 studies (76.9% vs. 74.6%, $p = .613$), open access versus paywalled articles (77.1% vs. 67.5%, $p = .176$), and highly versus less cited papers (74.4% vs. 78.1%, $p = .387$).

Moreover, when examining directional consistency regardless of statistical significance, GPT-4's responses aligned with the direction of human effects in 79.7% of main effects and 61.8% of interaction effects, both exceeding chance level (50%) at different significance levels ($p$ <.001 and $p$ = .011, respectively). This pattern suggests that even when strict replication criteria are not met, the model often captures the directional nature of human behavioral patterns.

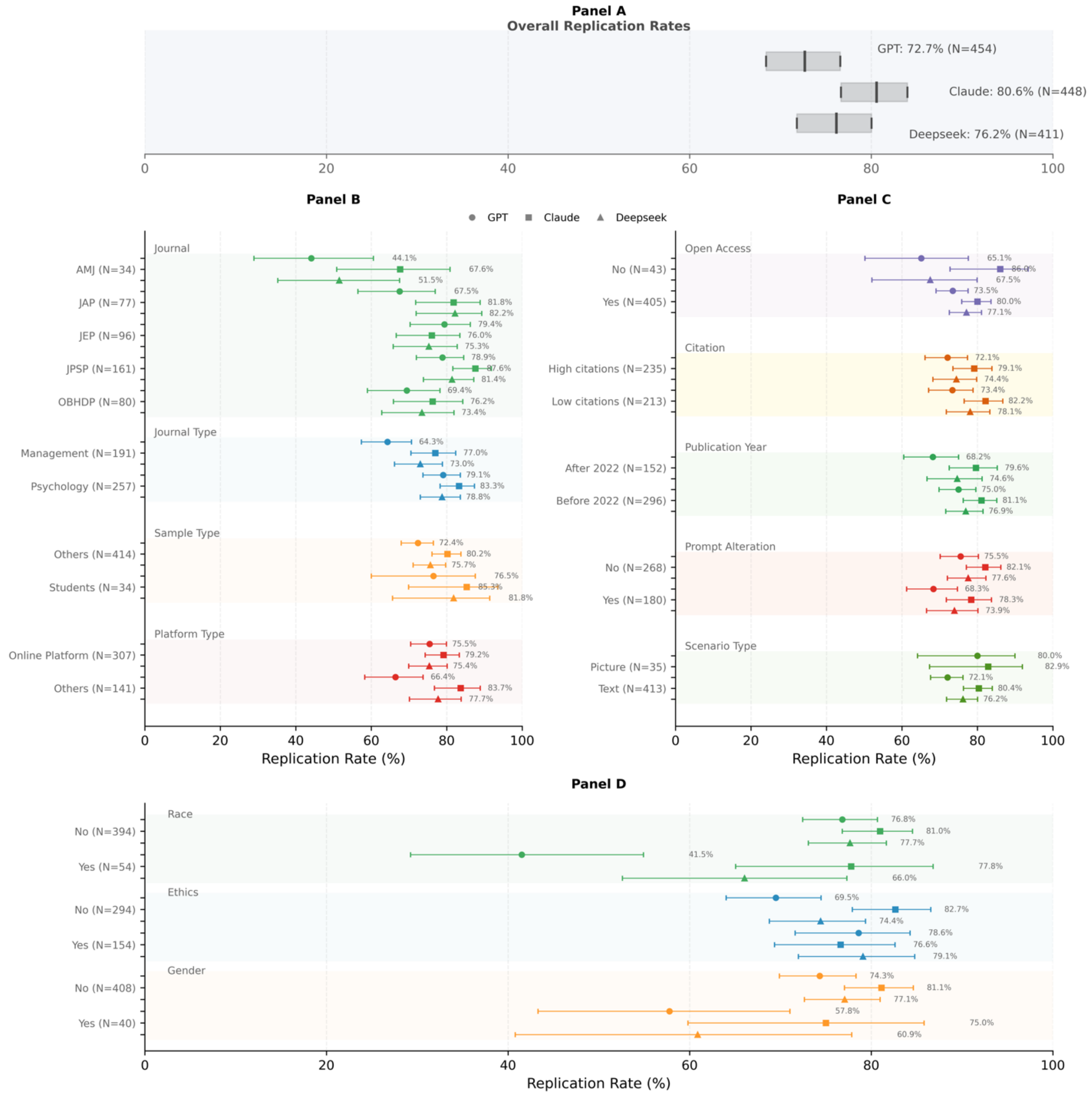

**Figure 2: Replication Rates of Main Effects across Study Characteristics**

Note: The figure is organized into four panels (A-D). Panel A shows the aggregate replication rate. Panels b-c present replication rates across various study attributes, where error bars indicate 95% confidence intervals. Journal abbreviations: AMJ (Academy of Management Journal), JAP (Journal of Applied Psychology), JEP (Journal of Experimental Psychology), JPSP (Journal of Personality and Social Psychology), OBHDP (Organizational Behavior and Human Decision Processes). "Race," "Ethics," and "Gender" categories indicate studies specifically investigating these social topics. Citation counts were dichotomized at the median into "Low citations" and "High citations." "Prompt alteration" refers to whether the original experimental prompts were modified. "Platform type" distinguishes between online crowdsourcing platforms and other participant recruitment methods.

## Statistical Significance Patterns

The distribution of p-values reveals distinct patterns between human studies and LLM replications. For main effects, all models produced significantly smaller p-values than human studies, with Claude showing the most pronounced difference using t-test ($M_{Claude} = 0.041$, $M_{human} = 0.112$, $p < .001$), followed by DeepSeek ($M_{DeepSeek} = 0.058$, $M_{human} = 0.113$, $p < .001$) and GPT-4 ($M_{GPT} = 0.078$, $M_{human} = 0.112$, $p = .004$). For interaction effects, the patterns varied markedly: Claude maintained significantly smaller p-values ($M_{Claude} = 0.048$, $M_{human} = 0.160$, $p < .001$), DeepSeek showed a similar trend ($M_{DeepSeek} = 0.097$, $M_{human} = 0.171$, $p = .046$), while GPT-4 showed no significant differences ($M_{GPT} = 0.178$, $M_{human} = 0.167$, $p = .758$).

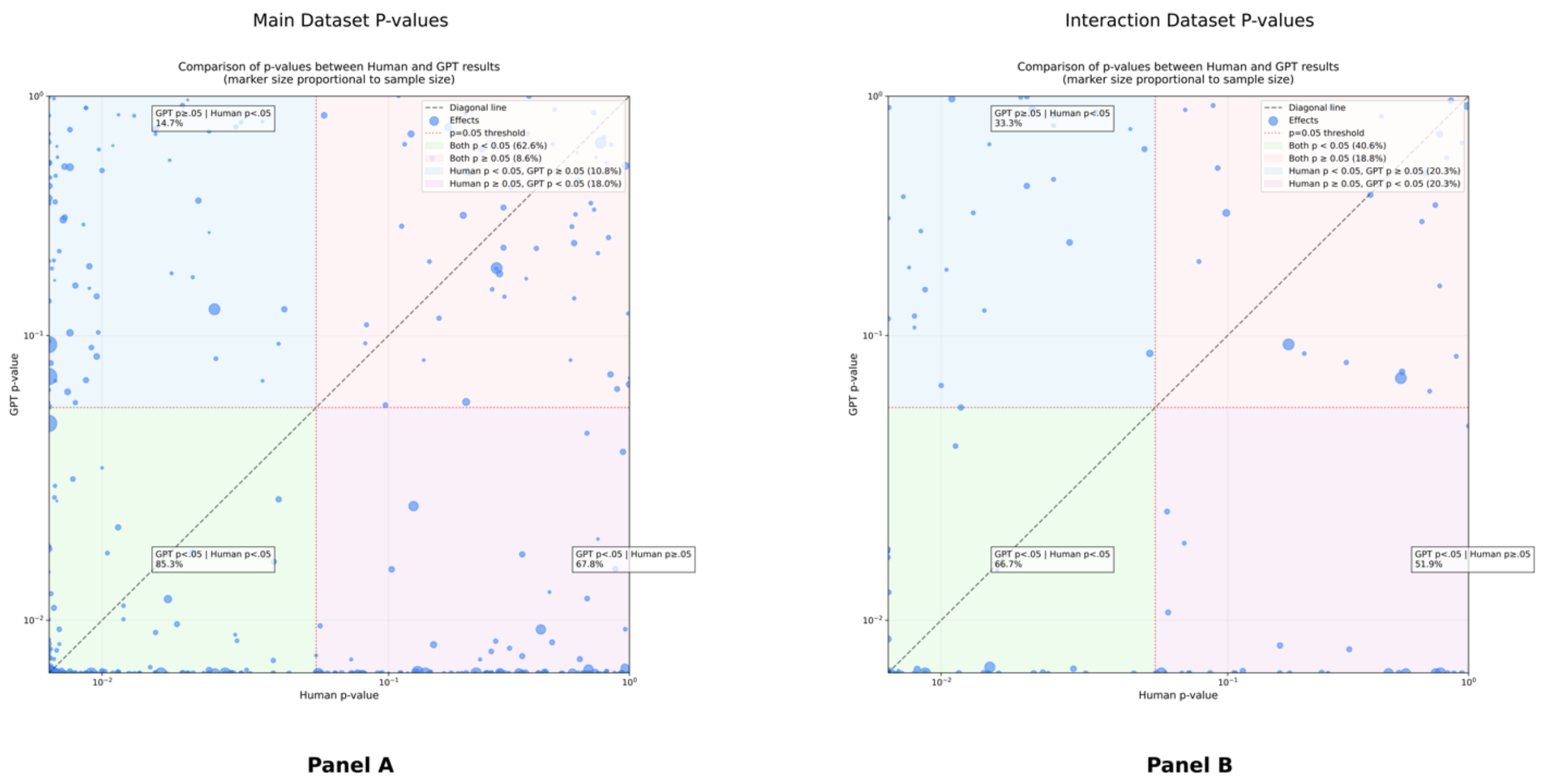

**Figure 3 Comparison for Original and Replication p-values of GPT-4**

Note: Visualization comparing p-values between original human studies (x-axis) and GPT replications (y-axis) across main effects (Panel A) and interaction effects (Panel B). Each scatter plot displays study pairs as dots whose size indicates sample size. Red threshold lines at $p = .05$ create quadrants showing different patterns of statistical significance agreement, with percentages indicating the proportion of studies in each category. Conditional probabilities (e.g., "GPT $p < .05$ | Human $p < .05$") quantify how often GPT replications maintain significance given the original study's significance status.

For main effects, all three language models demonstrated higher proportions of statistically significant findings ($\alpha = .05$) compared to human studies. However, the models diverge notably in their handling of interaction effects: while Claude and DeepSeek maintain elevated significance rates for both types of effects, GPT-4 shows a more nuanced pattern that more closely aligns with human studies for interactions. Figure 3 illustrates this pattern through scatter plots comparing p-values between human studies and GPT-4 replications (with Claude and DeepSeek results in Supplementary Materials). For main effects, chi-square tests revealed that GPT-4 shows higher significance rates than human studies (80.6% vs 73.4%, $p = .002$), consistent with other models (See SI Figure SI2-3). However, for interaction effects, GPT-4 uniquely shows no statistical difference in significance rates (60.9% vs 60.9%, $p = 1.000$), while both Claude and DeepSeek maintain significantly higher rates (87.4% vs 62.2%, $p < .001$ and 81.9% vs 62.9%, $p = .002$).

Importantly, we observed that AI models frequently produced significant findings when original human studies showed null effects ($p > .05$): for main effects, GPT-4 produced significant results in 67.8% of such cases, Claude in 79.5%, and DeepSeek in 82.8%, while for interaction effects, the rates were GPT-4 at 51.9%, Claude at 80.4%, and DeepSeek at 79.5% (Figure 3 and SI Figure SI2-3).

**Effect Size Analysis**

Our analysis focuses primarily on main effects for two key reasons. First, the smaller sample size for interaction effects reduces statistical reliability. More importantly, comparing effect sizes for interactions presents fundamental challenges, as the same interaction coefficient can produce qualitatively different patterns depending on the associated first-order terms. Therefore, while we present interaction results in the supplementary materials, our primary analysis concentrates on main effects where effect size comparisons are more straightforward.

For our main effects analysis, we retained samples where *r*-values from both the original studies and their corresponding replication studies were available and had clear directions, resulting in 578 cases. As shown in Figure 4-a, GPT-4 demonstrates a clear tendency towards larger effect sizes compared to the original studies ($M_{GPT} = 0.336$, $SD_{GPT} = 0.438$ vs. $M_{human} =$

0.246, $SD_{human}$ = 0.182; t = -5.604, *p* <.001). Similar patterns of effect size amplification are observed in the other language models, as detailed in Supplementary Materials.

Figure 4-b illustrates the relationship between original and replicated r-values for GPT-4, showing an overall Spearman correlation of 0.508(*p* <.001). Claude and DeepSeek show comparable correlations in supplementary analyses (ρ = 0.484, *p* <.001 and ρ = 0.483, *p* <.001). Across all models, psychology studies consistently show stronger correlations (GPT-4: ρ = 0.598, *p* <.001; Claude: ρ = 0.592, *p* <.001; DeepSeek: ρ = 0.591, *p* <.001) compared to management studies (GPT-4: ρ = 0.354, *p* <.001; Claude: ρ = 0.332, *p* <.001; DeepSeek: ρ = 0.333, *p* <.001), with these field differences being significant for each model (all *p* <.001). These correlations are comparable to previous replication efforts using human subjects (23), suggesting that while all three LLMs tend to produce larger effect sizes, their ability to maintain relative effect size relationships remains consistent with human-based replication studies.

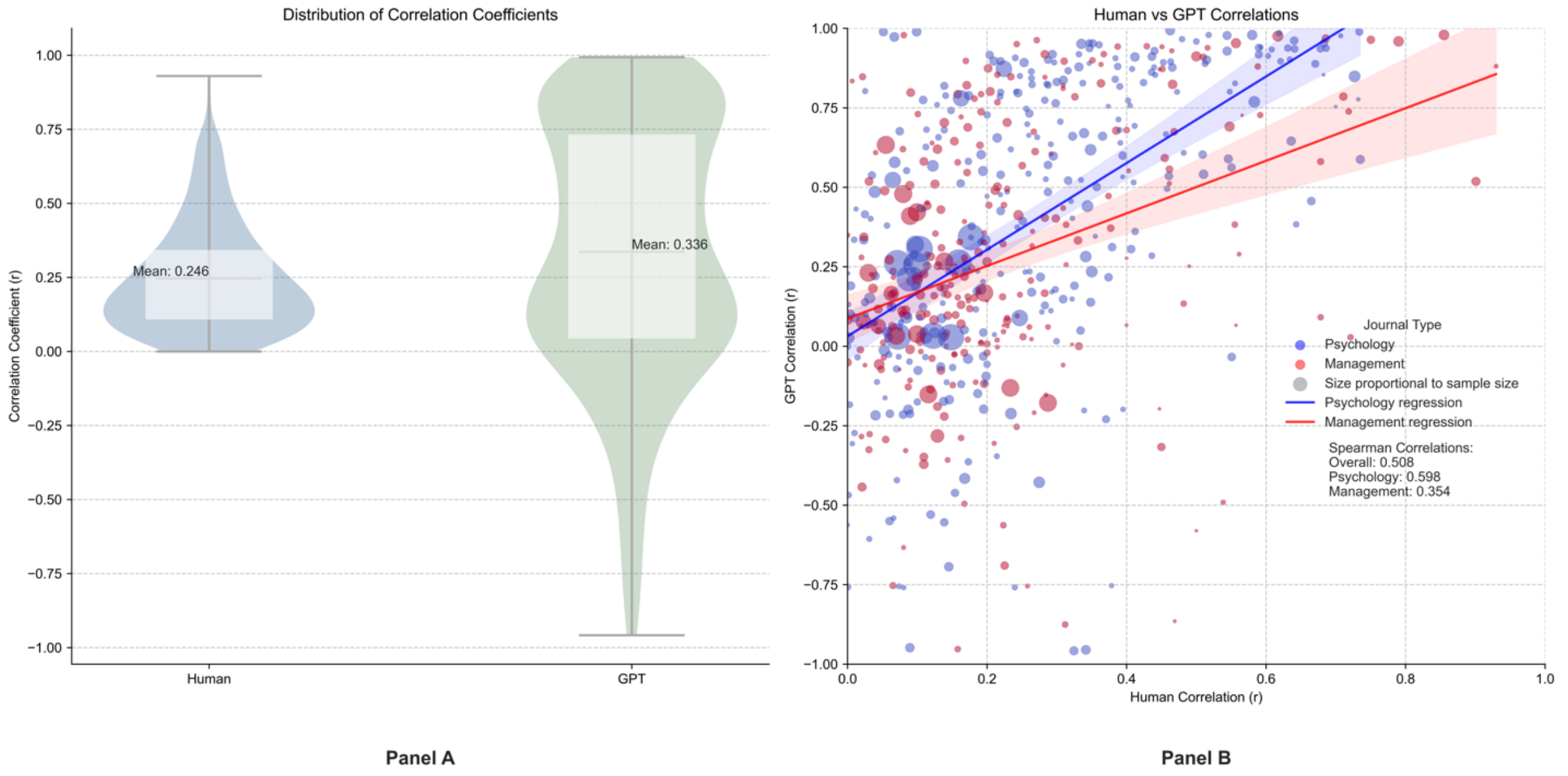

**Figure 4 Comparison for Original and Replication r-values of Main Effects**

Note: Comparison of correlation coefficients (r) between human and GPT studies. Panel A displays violin plots showing the distribution of correlation coefficients. Panel B plots human correlations (x-axis) against GPT correlations (y-axis), distinguishing between Psychology (blue) and Management (red) journals. Dot sizes represent sample sizes, with regression lines and confidence bands shown for each domain.

Our comparison of effect sizes between original and replicated studies revealed substantial systematic differences. For GPT-4, using Fisher's Z-transformation to properly account for uncertainty in both original and replicated correlations, we found that 65.92% of the effect pairs (381 out of 578) were significantly different from each other (see Figure 5-a). The mean Fisher's Z scores differed notably between human studies and all three language models, with human studies showing consistently lower values compared to GPT-4 ($M_{human}$ = 0.266, $M_{GPT}$ = 0.515; t = -8.744, *p* <.001), Claude ($M_{human}$ = 0.268, $M_{Claude}$ = 0.858; t = -13.147, *p* <.001), and DeepSeek ($M_{human}$ = 0.265, $M_{DeepSeek}$ = 0.634; t = -10.396, *p* <.001), indicating a consistent pattern of larger effect sizes in AI-generated results, with Claude showing the most pronounced amplification.

Further analysis of *r* values for GPT-4 revealed that 57.09% of cases showed significant differences between original and replicated studies (Figure 5-b). Notably, in 43.4% of cases, GPT-4's confidence intervals were positioned above those of the human studies (i.e., the lower bound of GPT-4's confidence interval exceeded the upper bound of the original human studies' confidence intervals), suggesting a systematic tendency toward larger effect sizes. Additionally, GPT-4 replications demonstrated consistently narrower confidence intervals in 76.1% of cases, with the mean CI width being significantly smaller for GPT-4 (M = 0.170) compared to human studies (M = 0.223; t = 10.648, *p* <.001). Similar patterns of higher positioning and narrower confidence intervals were observed for both Claude and DeepSeek replications (See SI Figure SI10, 12).

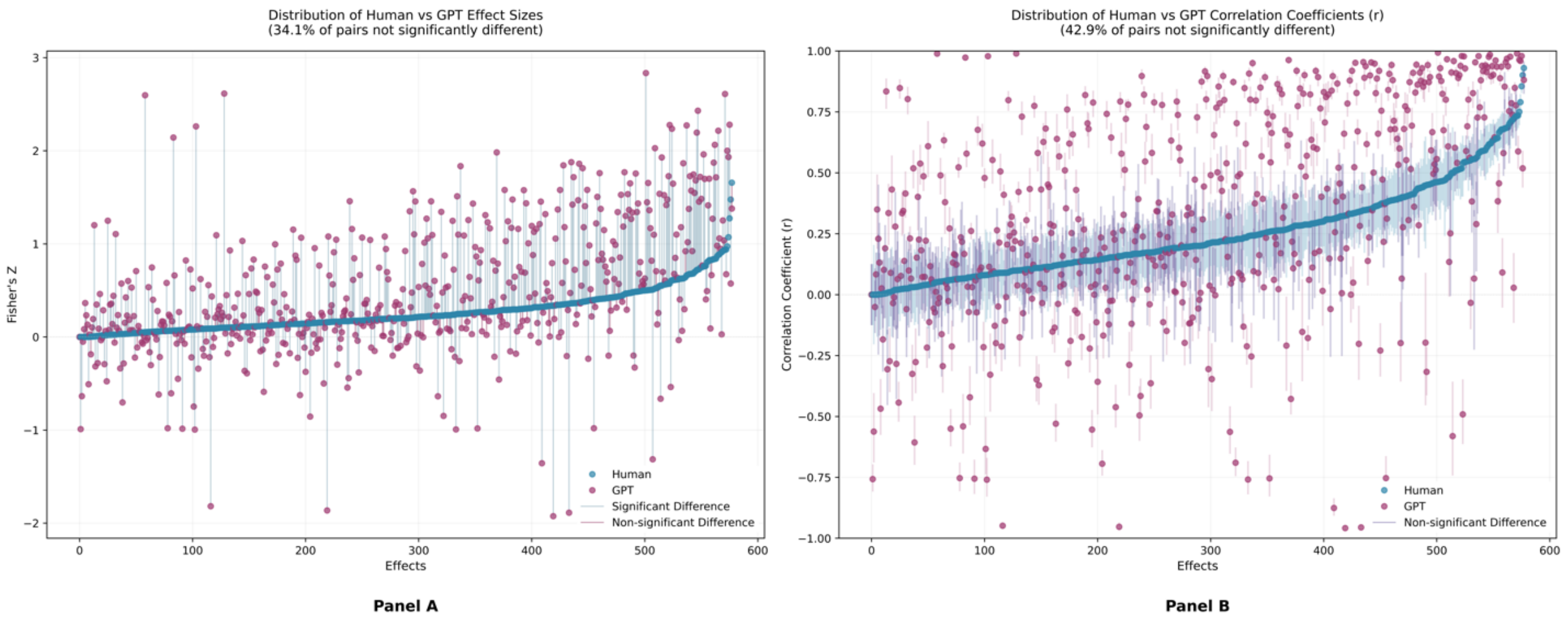

**Figure 5 Distribution of CI for r-values of Main Effects**

Note: Effect size comparison visualization between human and GPT studies. Panel A displays paired effect sizes using Fisher's Z transformation, with connecting lines between paired observations, with points color-coded to indicate statistical significance differences. From left to right, studies are ordered by original human effect size from smaller to larger, allowing for clear visualization of the relationship between effect magnitude and replication success. Panel B presents correlation coefficients with 95% confidence intervals for both human and GPT studies across the same dataset. Both panels utilize consistent color coding to distinguish between different statistical properties of the comparisons.

To better understand the mechanisms driving these larger effect sizes in LLM replications, we decomposed the contributions of between-group mean differences and within-group variance. For GPT-4, mean differences between experimental conditions increased by 147.3% compared to human studies (t = 3.685, *p* <.001, 95% CI = [0.687, 2.258]), while standard deviations decreased by 44.8% (t = -5.752, *p* <.001, 95% CI = [-0.601, -0.295]). Similar patterns emerged for Claude (mean differences: +271.0%, *p* <.001; standard deviations: -51.8%, *p* <.001) and DeepSeek (mean differences: +234.4%, *p* <.001; standard deviations: -44.2%, *p* <.001). These findings indicate that larger effect sizes in LLM replications are primarily driven by amplified between-group differences rather than reduced within-group variance, though both factors contribute.

It's worth noting that our study employed 1:1 sample size matching between original and LLM replications to isolate differences attributable solely to participant type while controlling for sample size. After discovering the pattern of larger LLM effect sizes, post-hoc power analyses revealed that our approach had sufficient power (>80%) to detect absolute differences in correlation coefficients (Δr) of 0.3 in most studies (95.1%), though many individual studies were underpowered for detecting smaller differences (0.1) or relative differences (30%). The fact that we consistently observed larger LLM effect sizes despite this conservative testing approach strengthens our conclusions and suggests the phenomenon is robust. All data and analysis scripts are available in the OSF repository.

**Antecedents of Replication Rate and Effect Size Distribution**

To examine the factors that significantly influence replication success (1 = replicated, 0 = not replicated), the difference between effect sizes (GPT r - human r), and the consistency between the directions (1 = consistent, 0 = not consistent), we conducted regression analyses (Table 1).

Significant findings revealed that studies involving race or ethnicity (b = -1.873, *p* <.001) and gender variables (b = -1.060, *p* = .003) had lower replication success rates, indicating challenges in replicating effects related to these socially sensitive topics. Additionally, experiments that required scenario adaptations for GPT-4 (b = 0.100, $p = .736$) showed a positive but not significant impact on replication success. Even with modifications to make the scenarios more accessible to the model, the complexity or nuance of these experiments likely resulted in no significant change in replication success.

When examining effect size differences (GPT r - human r), race-related variables showed significantly smaller deviations from original human studies (b = -0.220, $p = .006$), while ethical and moral variables did not (b = -0.047, $p = .179$). The analysis did not show that studies involving variables related to race or ethnicity (b = -0.670, $p = .125$) were significantly more likely to generate inconsistent effect directions in replication. Nonetheless, gender topic (b = -0.624, $p = .036$) significantly led to direction inconsistency, while larger original effect sizes (b = 4.941, $p <.001$) significantly led to direction consistency.

Parallel regression analyses for Claude and DeepSeek revealed distinct patterns of predictor relationships, reflecting the unique characteristics of each model observed earlier in Figure 2. While sharing some commonalities with GPT-4, each model showed different sensitivities to various study characteristics, particularly in their handling of social variables. Detailed regression results for Claude and DeepSeek, along with comparative analyses, are provided in the supplementary materials (see Table SI2 and SI3).

**Table 1: Predictors of Replication Rate and Effect Size Distribution of GPT Main Effects**

| | (1) | (2) | (3) |
|---|---|---|---|
| **Regression Analysis I** | | | |
| | **Replication success** | **Effect size difference** | **Direction Consistency** |
| Management Journal | -0.488* | -0.039 | -0.066 |
| | (0.243) | (0.036) | (0.222) |
| Online Platform | -0.198 | -0.059 | -0.090 |
| | (0.288) | (0.043) | (0.271) |
| Gender Topic | -1.060** | -0.093 | -0.624* |
| | (0.352) | (0.052) | (0.298) |
| Race Topic | -1.873*** | -0.220** | -0.670 |
| | (0.461) | (0.080) | (0.437) |
| Social Relationships Topic | -0.217 | -0.038 | 0.198 |
| | (0.286) | (0.036) | (0.250) |
| Ethics Topic | 0.116 | -0.047 | 0.058 |
| | (0.274) | (0.035) | (0.234) |
| Emotion Topic | -0.647 | 0.039 | -0.298 |
| | (0.360) | (0.060) | (0.324) |
| Technology Topic | -0.484 | 0.056 | 1.199 |
| | (0.636) | (0.095) | (1.069) |
| Prompt Alteration | 0.100 | -0.028 | -0.199 |
| | (0.296) | (0.038) | (0.243) |
| _cons | 1.825*** | 0.211*** | 1.649*** |
| | (0.338) | (0.041) | (0.302) |
| N | 454 | 578 | 606 |
| BIC | 548.276 | 578.502 | 660.578 |
| Log pseudolikelihood | -243.547 | -257.453 | -298.255 |
| **Regression Analysis II** | | | |
| | (1) | (2) | (3) |
| | **Replication success** | **Effect size difference** | **Direction Consistency** |
| Original Effect Size | 4.328*** | 0.142 | 4.941*** |
| | (0.901) | (0.080) | (0.845) |
| _cons | -0.165 | 0.055* | 0.378* |
| | (0.253) | (0.025) | (0.179) |
| N | 434 | 578 | 578 |
| BIC | 480.468 | 550.066 | 546.303 |
| Log pseudolikelihood | -234.161 | -268.673 | -266.792 |

Note: $^{*}p$ <.05; $^{**}p$ <.01; $^{***}p$ <.001. The variable "Management Journal" is coded as 1 for management journals (AMJ, JAP, OBHDP) and 0 for psychology journals (JEP, JPSP). The variable "Online Platform" is coded as 1 for studies conducted on MTurk or Prolific platforms, and 0 for other platforms. "Gender Topic" refers to variables related to gender, while "Race Topic" pertains to variables related to race and ethnicity, including race, country, etc. "Social Relationships Topic" includes variables related to social hierarchy and relationships, such as power, status, compliance, justice, norms, inequality, corruption, hierarchy, etc. "Ethics Topic" covers variables related to ethical and moral issues, including mistreatment, moral objections, unethical behavior, etc. "Emotion Topic" includes variables related to human emotions, such as passion, respect, liking, warmth, anxiety, pride, etc. "Technology Topic" refers to variables related to technology, including algorithms. Lastly, "Prompt Alteration" is coded as 1 when adaptation was made to the scenario, and 0 when no adaptation was necessary. All variables were entered into the regression model simultaneously, except for Original Effect Size, as its sample differs from that of the other variables. DV1 and DV3 are binary, thus logistic regression was used. DV2 was analyzed using ordinary least squares (OLS) regression.

**Analysis of Temperature Effects on Replication Outcomes**

We initially used default temperature settings recommended by each LLM to balance variability and precision in responses, maintaining standard operating conditions. To comprehensively examine temperature's influence on replication outcomes, we used GPT-4 as an illustrative example, focusing on a strategically selected subset of studies for feasibility. Specifically, we identified studies containing at least one borderline case (where GPT p-values fell between .05 and .10), as these marginally significant results provided ideal test cases for examining whether temperature adjustments could systematically shift significance patterns. We analyzed all effects from these selected studies across three temperature settings (0, 0.5, and 1.0), examining all 64 effects including 11 borderline cases to capture both temperature sensitivity in marginal results and broader patterns across the full dataset.

Statistical analyses revealed that temperature settings had surprisingly little systematic effect on results. Contrary to expectations, lower temperatures did not consistently lead to higher statistical significance rates ($\alpha = .05$) (Figure 6, Panel A). Effect sizes remained remarkably stable across temperatures (means: 0.172 at temp 0, 0.154 at temp 0.5, 0.161 at temp 1.0;

ANOVA F = 0.030, $p$ = .971), with GPT r-values and p-values showing similar consistency across temperature settings (Panel C). As shown in Panel B, replication rates modestly increased from temperature 0 to 1.0, though these differences were not statistically significant (see the OSF repository). These findings suggest that temperature adjustments had minimal impact on effect magnitudes or significance patterns in our analyses.

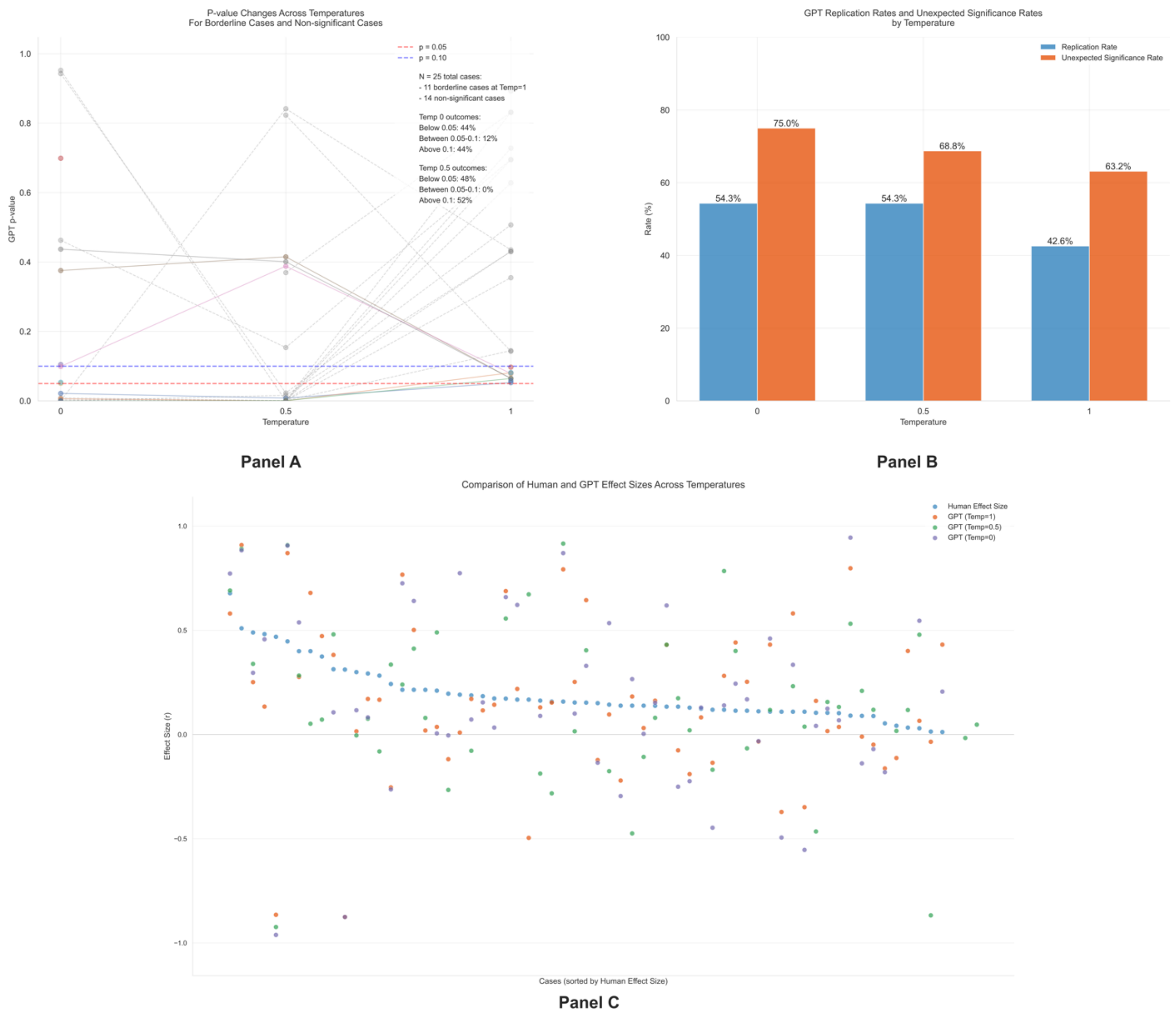

**Figure 6: Temperature Effects on Replication Outcomes in Selected GPT-4 Studies**

Note: Temperature-dependent consistency in GPT-4's statistical replication results. This figure examines how varying GPT-4's temperature parameter affects its statistical judgments, focusing on studies containing borderline significant cases (where GPT-4 at temperature=1 found $0.05 \leq p \leq 0.1$). Panel A tracks how p-values of

individual effects change across temperature settings (0, 0.5, 1), with solid and dashed lines distinguishing between initially significant and non-significant results. Panel B presents both the successful replication rates of originally significant effects and the rates of unexpected significant results (where GPT-4 found significance in originally non-significant cases) across different temperature settings. Panel C compares the magnitude of replicated effect sizes across different temperatures against the original effect sizes. Red and blue dashed lines mark conventional significance thresholds ($p = .05$ and $p = .10$, respectively).

## Discussion

Our study provides a comprehensive examination of LLMs' capability to replicate randomly selected experiments from top management and psychology journals through simulated participant responses. The results reveal systematic differences between LLM and human studies, particularly for main effects, with LLMs showing consistently larger effect sizes, narrower confidence intervals, and a strong tendency toward statistical significance. This pattern varies across different types of studies and experimental designs.

### Implications for Experimental Social Science with LLMs

These findings both align with and diverge from recent work examining LLMs in social science research. While studies like Ashokkumar et al. (2024) found high correlations between LLM-predicted and human effect sizes (r > .90), our results suggest that the reliability of LLM-based replications may depend heavily on the nature of the studies being examined. The discrepancy likely stems from some fundamental differences in the types of research being replicated. Ashokkumar et al. focused primarily on large-scale intervention studies with straightforward directional effects and practical behavioral outcomes. In contrast, our sample drew from top-tier psychology and management journals that often prioritize novel, even counterintuitive findings and complex experimental designs - characteristics that may make these effects inherently more challenging to replicate, whether by humans or AI. This pattern mirrors broader challenges in psychological science, as highlighted by the replication crisis where large-scale efforts found that complex, counterintuitive findings tend to be particularly difficult to reproduce in human studies (28).

Our methodological approach advances beyond previous work by providing a more comprehensive assessment of LLMs' capabilities in experimental research. While Lippert et al. (2024) demonstrated LLMs' potential for predicting overall experimental outcomes, we expanded this line of inquiry by examining psychological phenomena across diverse experimental paradigms. Our key innovation was simulating individual participant responses for each experimental condition, allowing us to assess whether LLMs can capture the underlying psychological mechanisms that drive human behavior and decision-making.

The systematic tendency of LLMs to produce stronger statistical relationships (as evidenced by larger effect sizes and higher proportions of statistically significant results at $\alpha = .05$) compared to human studies may represent an advantage in certain contexts. Like ideal research participants, LLMs remain consistently focused and are free from the noise introduced by fatigue, distraction, or varying attention levels that characterize human subjects. This reduced heterogeneity in responses makes LLMs particularly valuable for pilot studies and initial tests of experimental instruments, especially given their reliable replication of effect directions even when effect magnitudes differ from human studies (3,8).

Beyond pilot testing, LLMs offer unprecedented opportunities for rapid hypothesis testing and iterative refinement of experimental designs before committing to resource-intensive human studies. Their ability to process large numbers of experimental conditions quickly and cost-effectively enables researchers to explore broader parameter spaces and identify promising research directions. When integrated with traditional methods, this scalability could accelerate the research cycle while preserving resources for the most promising lines of inquiry, potentially opening new avenues for psychological research that were previously impractical to pursue.

However, these characteristics also necessitate careful interpretation. While increased statistical power and cleaner signals might seem advantageous, they may lead to overestimation of real-world effect sizes and potentially misleading conclusions about practical significance. This concern is particularly acute for studies involving sensitive topics like gender or race, where LLMs may reflect and amplify existing societal biases. The comparison between management and psychology journal replications is particularly telling - lower replication rates for management studies, which traditionally favor novel and counterintuitive findings, suggest that LLMs may be less effective at capturing effects that arise from complex, context-dependent mechanisms or those that may have been influenced by publication bias and questionable research practices.

These findings raise important questions about using published research as the benchmark for evaluating LLM performance. In the context of the replication crisis and increasing awareness of publication bias, there may be cases where discrepancies between LLM and published results reflect limitations in the original studies rather than shortcomings of the AI models. This underscores the potential value of LLMs as tools for methodological triangulation while highlighting the importance of maintaining a nuanced perspective on both human and AI-generated research findings.

Our analysis of interaction effects further illuminates the varying capabilities and limitations across different LLMs in experimental research. While GPT-4 showed moderate success in replicating interaction effects without the inflation bias seen in main effects, Claude and DeepSeek demonstrated different patterns in their handling of interactive mechanisms. This variation across models in capturing interaction effects—which typically require more sophisticated understanding of how variables work together—suggests different levels of capability in processing complex psychological relationships.

Our analysis also revealed instances (1-2.5% of effects) where LLMs produced completely uniform responses across all simulated participants, making effect size calculations mathematically impossible due to zero variance. While GPT-4 exhibited this primarily in ethical scenarios, Claude and DeepSeek showed similar patterns across broader contexts. These "blind spots," where LLMs fail to simulate natural human response variability, represent an important constraint for researchers using these models in experimental simulations and warrant consideration in future methodological approaches.

Another notable limitation of our study is its GPT-centric approach to prompt engineering. While our prompts were systematically developed and refined through pretesting with GPT-4, we did not conduct equivalent prompt optimization processes for Claude and DeepSeek. Consequently, our cross-model comparisons should be interpreted primarily as robustness checks rather than definitive performance benchmarks, with Claude and DeepSeek results serving to validate the generalizability of our core findings rather than establish absolute performance rankings across models.

**Implications for Understanding LLMs and Human Cognition**

Our study not only sheds light on the replicability of psychological experiments but also offers insights into the behavior of LLMs themselves. By using a range of psychological experiments as benchmarks, we can better understand how LLMs process information and respond to various stimuli.

The comparison between LLM and human responses to these experiments provides a nuanced understanding of where AI and human cognition converge and diverge (4,11–13,16,41). This is crucial as we progress towards more advanced AI systems and potentially artificial general intelligence (AGI). Moreover, this comparative approach between LLMs and human responses in psychological experiments offers a unique window into the inherent biases and limitations of AI systems (17–19). As LLMs are increasingly integrated into decision-making processes across various sectors of society, understanding these biases becomes crucial (42). Notably, our findings reveal that different LLMs exhibit distinct patterns in handling socially sensitive topics such as race, gender, and ethics, suggesting varying degrees of alignment with human values and social norms. These systematic differences between models in processing sensitive social content provide valuable insights into how different training approaches and safety measures may influence an AI system's alignment with human values and societal expectations.

While our study represents a substantial replication effort encompassing over 150 studies - a notably large sample compared to previous replication initiatives – it's important to acknowledge that this still represents only a fraction of potential candidate studies suitable for LLM-based replication. Our approach balanced scalability with feasibility, but this inherently introduces potential sampling biases that warrant consideration. This limitation is particularly salient when examining interaction effects, where our smaller sample size necessitates more cautious interpretation of the findings. We observed various patterns in the alignment between LLM and human responses, including both convergent and divergent cases. However, the relatively few instances of divergence merit further investigation. Future research should explicitly focus on these cases of divergence to better understand whether they stem from limitations in LLM capabilities, inherent challenges in human study reproducibility, or other factors. Such targeted investigation could provide valuable insights into both the capabilities and limitations of LLMs as tools for psychological research, while helping to establish more precise boundaries for their application in experimental replication.

In conclusion, while LLMs are not yet a replacement for human-subject research, they offer a powerful and cost-effective tool for preliminary hypothesis testing, experimental design refinement, and exploring the broader implications of psychological theories. As we continue to explore and improve the capabilities of LLMs, these models may open new avenues for interdisciplinary research, merging insights from social science, computer science, and artificial intelligence. Such research could lead to a deeper understanding of both human cognition and the evolving role of AI in scientific inquiry.